\documentclass{article} 
\usepackage{iclr2026_conference,times}

\usepackage{amsmath,amsfonts,bm}

\def\eqref#1{equation~\ref{#1}}

\def\1{\bm{1}}

\DeclareMathAlphabet{\mathsfit}{\encodingdefault}{\sfdefault}{m}{sl}
\SetMathAlphabet{\mathsfit}{bold}{\encodingdefault}{\sfdefault}{bx}{n}

\usepackage{url}
\usepackage{graphicx}
\usepackage{booktabs}
\usepackage{multirow}
\usepackage{xcolor}
\usepackage{algorithm}
\usepackage{algpseudocode}
\usepackage{hyperref}

\title{DeLS-Spec: Decoupled Long-Short Contexts for Parallel Speculative Drafting}

\author{Hong-Kai Zheng, Piji Li
\thanks{Corresponding author} \\
College of Artificial Intelligence,\\
Nanjing University of Aeronautics and Astronautics, China\\
MIIT Key Laboratory of Pattern Analysis and Machine Intelligence, Nanjing, China\\
The Key Laboratory of Brain-Machine Intelligence Technology, \\Ministry of Education, Nanjing, China\\
\texttt{\{dt\_ttt,pjli\}@nuaa.edu.cn}
}

\newcommand{\tabsmall}{\scriptsize}
\newcommand{\secondarycolor}[1]{\textcolor{gray}{#1}}

\iclrpreprintcopy 

\begin{document}

\maketitle

\begin{abstract}
Speculative decoding accelerates LLM inference by drafting multiple tokens and verifying them in parallel. Block-parallel drafters such as DFlash further improve drafting efficiency by predicting an entire block in one pass, but their position-wise predictions lack explicit intra-block causal conditioning. Recent methods such as Domino and DSpark attempt to introduce such causality into block-parallel drafting, but they require training the draft model from scratch, which limits their flexibility and increases training cost.
We propose \textbf{DeLS-Spec}, a decoupled long-short context speculative decoding method. DeLS-Spec treats the fixed DFlash model as a long-context expert and introduces a lightweight local head as a short-context expert. The local head can be trained independently with a standard next-token prediction objective, without joint training with the target model or the DFlash backbone, leading to extremely low training cost. At inference time, DeLS-Spec combines long-context and short-context logits, and the local head is not tied to a specific DFlash checkpoint, making the method more modular and flexible. Experiments on Qwen3 models show that DeLS-Spec consistently improves speedup and average acceptance length over DFlash across math, code, and dialogue benchmarks. 
Code is available at \href{https://github.com/dt-3t/DeLS-Spec}{GitHub}.
\end{abstract}

\section{Introduction}

Large language models are typically decoded 
autoregressively \citep{achiam2023gpt,liu2024deepseek,yang2025qwen3}, 
where each generated token requires a new forward pass of the 
target model. This sequential process becomes a major latency 
bottleneck, especially for long-form generation and interactive 
applications. Speculative decoding mitigates this problem by 
using a lightweight draft model to propose multiple future 
tokens, which are then verified in parallel by the target model 
while preserving the target 
distribution \citep{leviathan2023fast,chen2023accelerating}.

The efficiency of speculative decoding depends heavily on 
the drafter \citep{cai2024medusa,ankner2024hydra,li2024eagle,li2024eagle2,li2026eagle}. Traditional autoregressive drafters maintain 
strong local consistency but still generate draft tokens 
sequentially \citep{li2024eagle,cheng2024recurrent}. Recent block-parallel drafters, such as 
DFlash, improve drafting efficiency by predicting an entire 
block of tokens in one pass \citep{an2025pard,liu2026dart,chen2026dflash}. 
However, this parallelism also introduces a limitation: 
tokens inside the same draft block are predicted largely 
independently, without explicitly conditioning on previously 
drafted tokens in that block \citep{chen2026dflash,huang2026domino,rheinboldt2026treeflash}. As a result, DFlash provides 
strong long-context predictions from the prefix, but lacks 
explicit short-context causal modeling within the draft block.

Several recent methods, including Domino and DSpark, attempt to 
address this issue by introducing intra-block causality into 
block-parallel drafting \citep{huang2026domino,dspark,hu2026jetspecbreakingscalingceiling,rheinboldt2026treeflash}. These 
methods show that modeling causal dependencies inside the draft 
block can improve acceptance. However, they usually require 
training a new draft model from scratch or jointly training the 
draft backbone with additional causal components. This makes them 
costly to apply when a DFlash-style drafter has already been 
trained, and limits their flexibility across different draft checkpoints.

In this work, we propose \textbf{DeLS-Spec}, a decoupled 
long-short context speculative decoding method. Instead of 
modifying or retraining the DFlash backbone \citep{chen2026dflash}, DeLS-Spec keeps 
DFlash fixed as a long-context expert and introduces a lightweight 
local head as a short-context expert. The DFlash expert captures 
semantic and task-level information from the full prefix, while 
the local head captures causal dependencies from the already drafted 
tokens inside the current block.

A key property of DeLS-Spec is its decoupled training. The local 
head is trained independently with a standard next-token prediction 
objective on plain text data. It does not require target-model hidden 
states, DFlash hidden states, or joint optimization with the speculative 
decoding pipeline. This makes training extremely cheap and allows the 
local head to be attached to existing DFlash-style checkpoints after 
they have been trained. At inference time, DeLS-Spec combines the long-context logits from 
DFlash and the short-context logits from the local head \citep{hinton2002training}. Since the 
local head is not trained together with a specific DFlash checkpoint, 
it is not one-to-one tied to the DFlash weights during inference. 
This makes the method more modular and flexible: a trained local head 
can be reused or transferred across compatible DFlash checkpoints, 
rather than requiring a new draft model to be trained for each setting.

We evaluate DeLS-Spec on Qwen3 models \citep{yang2025qwen3} across math, code, 
and dialogue benchmarks. Experiments show that DeLS-Spec consistently 
improves both decoding speedup and average acceptance length over DFlash. 
These results demonstrate that decoupling long-context block-parallel 
drafting from short-context causal correction is an efficient and 
practical way to enhance existing speculative decoding systems.

Our contributions are summarized as follows:
\begin{itemize}
\item We propose DeLS-Spec, a decoupled long-short context speculative decoding method that improves DFlash-style block-parallel drafting without retraining the DFlash backbone.
\item We introduce a lightweight local head that provides intra-block causal information and can be trained independently with a standard next-token prediction objective.
\item DeLS-Spec consistently enhances DFlash on Qwen3 models across math, code, and dialogue benchmarks, while offering modular flexibility across different checkpoints.
\end{itemize}

\section{Related Work}
\label{sec:related_work}

\paragraph{Speculative decoding and classical drafters.}
Speculative decoding accelerates autoregressive inference by using a lightweight drafter to propose multiple tokens and a target model to verify them in parallel while preserving the target distribution \citep{leviathan2023fast,chen2023accelerating}. Early blockwise decoding and tree-based verification reduce sequential decoding rounds or expand candidate coverage \citep{stern2018blockwise,miao2024specinfer}. Later drafters improve proposal quality through target-attached heads, sequentially dependent heads, feature-level drafting, dynamic draft trees, recurrent drafting, and distillation \citep{cai2024medusa,ankner2024hydra,li2024eagle,li2024eagle2,li2026eagle,cheng2024recurrent,zhou2024distillspec}. 
However, many drafters still require sequential drafting, feature updates, or tree construction, making drafting cost grow with the speculative budget.

\paragraph{Parallel and diffusion-based drafting.}
Parallel generation reduces drafting latency by predicting multiple positions simultaneously, but also weakens token dependency, as observed in non-autoregressive and diffusion-style language generation \citep{gu2017non,nie2026large,arriola2025block,cheng2026sdar,liu2025tidar}. Recent speculative drafters therefore use diffusion or parallel prediction to generate draft blocks efficiently \citep{christopher2025speculative,li2026diffuspec,sandler2025specdiff,an2025pard,liu2026dart}. DFlash is most relevant to our work: it uses a lightweight block-diffusion drafter with target hidden-state injection to obtain strong long-context conditioning and generate a whole block in one pass \citep{chen2026dflash}. Follow-up methods build draft trees or improve training objectives for such parallel drafters \citep{ringel2026accelerating,wu2026d}. Nevertheless, DFlash-style block-parallel prediction mainly models position-wise long-context distributions and lacks explicit conditioning on the locally drafted prefix, which can hurt later-token acceptance.

\paragraph{Intra-block causality and parallel-drafter correction.}
Several methods address the missing local causality in parallel drafting. 
Hydra introduces dependency among draft heads \citep{ankner2024hydra}; 
Domino corrects DFlash-style logits with a GRU causal encoder and low-rank 
residual head \citep{huang2026domino}; DSpark adds semi-autoregressive 
Markov/RNN heads and confidence scheduling \citep{dspark}; 
JetSpec uses tree-causal attention for branch-wise causal 
conditioning \citep{hu2026jetspecbreakingscalingceiling}; 
and TreeFlash approximates local autoregressive distributions on top of 
DFlash \citep{rheinboldt2026treeflash}. D-PACE further shows that accepted 
length is tied to position-dependent prefix 
acceptance \citep{wu2026d}. Although structured non-autoregressive 
decoding and CTC-based drafting can introduce dependencies or 
alignments \citep{sun2019fast,wen2024speculative}, they are less 
compatible with the exact per-position probabilities required 
by standard lossless verification. Unlike these unified correction 
pipelines, DeLS-Spec keeps the trained DFlash drafter fixed as a 
long-context expert, independently trains a lightweight short-context 
expert for intra-block causality, and fuses their logits with unigram-prior 
correction to avoid double-counting frequency bias.

\section{Preliminaries} 
\label{sec:preliminaries}

\subsection{Parallel Draft Model: DFlash} 
\label{sec:prelim_dflash}

Let $y=x_{<k}$ denote the observed long context, and let
$x_k,x_{k+1},\ldots,x_{k+s}$ denote the block of future tokens to be drafted.
Parallel draft models such as DFlash \citep{chen2026dflash} perform block-parallel drafting by
predicting multiple future tokens in a single forward pass. Specifically, for
each position inside the draft block, the model predicts
\begin{equation}
    p_L(x_i \mid y),
    \qquad i\in\{k,\ldots,k+s\}.
\end{equation}
Although DFlash is described as a diffusion language model, its practical
drafting procedure consists of one forward pass followed by one sampling step,
which can be viewed as a diffusion language model with a single denoising step.
This design significantly reduces draft latency compared with autoregressive
draft models. However, because tokens within the block are predicted in
parallel, each position is unaware of the tokens generated at other positions.
As a result, the causal dependency inside the draft block is weak, which limits
the accepted length during speculative decoding.

\subsection{Introducing Intra-Block Causality} 
\label{sec:prelim_causal}

Recent approaches such as Domino \citep{huang2026domino} and DSpark \citep{dspark} address this issue by introducing
intra-block causal modeling on top of the parallel DFlash backbone. They attach
a lightweight causal correction head after the parallel backbone and directly
optimize
\begin{equation}
    \mathcal{L}_{\mathrm{cc}}
    =
    -\sum_{i=k+1}^{k+s}
    \log p(x_i \mid y,z_i),
    \qquad
    z_i=x_{k:i-1}.
\end{equation}
where $z_i$ denotes the local prefix within the draft block. Although this
causal correction improves local consistency, it is typically trained as part
of the complete draft pipeline. In particular, it depends on the target model
and requires joint training or fine-tuning of the parallel DFlash backbone and
the causal correction head. Therefore, migrating such causal dependency
mechanisms to an already trained DFlash model can be expensive, especially when
the model size increases.

\section{Method}
\label{sec:method}

\begin{figure}[t]
    \centering
    \includegraphics[width=0.98\linewidth]{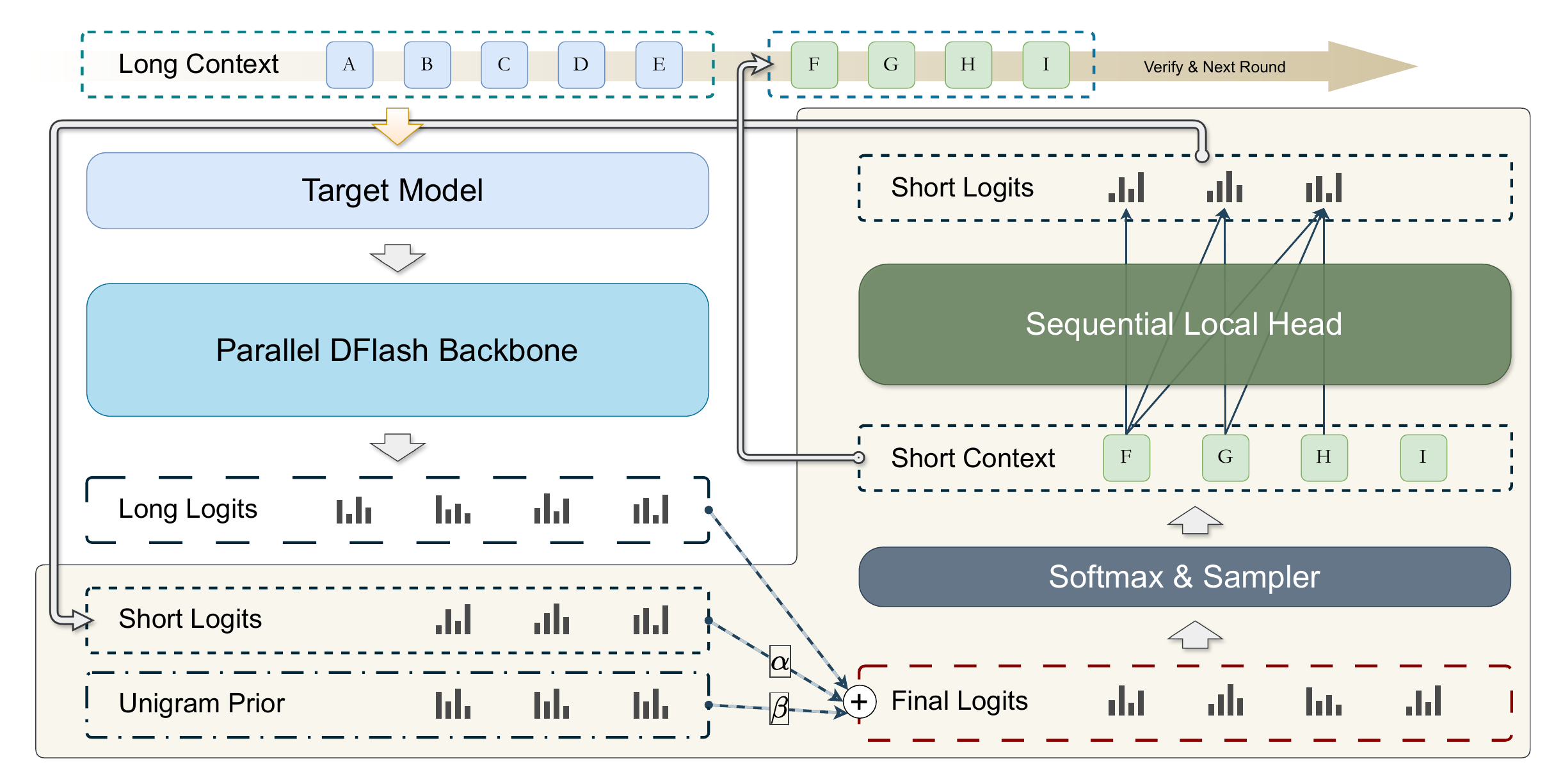}
\caption{
Overview of DeLS-Spec. DFlash produces long-context logits in parallel, while a
lightweight local head sequentially estimates short-context logits from the
draft prefix. DeLS-Spec fuses them with a unigram prior to obtain final draft
logits, which are sampled and then verified by the target model under standard
speculative decoding.
}
    \label{fig:main-overview}
\end{figure}

\subsection{Decoupled Long-Short Contexts}
\label{sec:dels}

We propose \textbf{DeLS-Spec}, a decoupled long-short context modeling method
for speculative drafting. The key observation is that an already trained DFlash
model provides a long-context conditional distribution $p_L(x_i\mid y)$, which
captures semantic and task-level constraints from the full prefix $y$. What is
missing is the short-range causal dependency induced by the local block prefix
$z_i$. Therefore, a natural starting point is a product-of-experts \citep{hinton2002training} formulation:
\begin{equation}
    p(x_i \mid y,z_i)
    \approx
    p(x_i \mid y)\,p(x_i \mid z_i).
\end{equation}
In this view, the long-context expert provides global semantic constraints,
while the short-context expert provides local coherence constraints within the
draft block. However, this factorization is not an exact equality. More
precisely, we have
\begin{equation}
\begin{aligned}
    p(x_i \mid y,z_i)
    &=
    \frac{
        p(x_i \mid y)\,p(x_i \mid z_i)
    }{
        p(x_i)
    }
    \cdot
    \frac{
        p(y)\,p(z_i)
    }{
        p(y,z_i)
    }
    \cdot
    \frac{
        p(y,z_i \mid x_i)
    }{
        p(y \mid x_i)\,p(z_i \mid x_i)
    } .
\end{aligned}
\end{equation}
The proof of this factorization is provided in Appendix~\ref{app:proof_factorization}.
The second factor,
\begin{equation}
    \frac{p(y)\,p(z_i)}{p(y,z_i)},
\end{equation}
does not depend on the candidate token $x_i$ and can therefore be absorbed into
the normalization constant when applying softmax or ranking candidate tokens.
Thus, the token-dependent part can be written as
\begin{equation}
    p(x_i\mid y,z_i)
    \propto
    \frac{
        p(x_i \mid y)\,p(x_i \mid z_i)
    }{
        p(x_i)
    }
    \cdot
    \frac{
        p(y,z_i \mid x_i)
    }{
        p(y \mid x_i)\,p(z_i \mid x_i)
    } .
\end{equation}

This decomposition yields two important implications. First, directly
multiplying $p(x_i\mid y)$ and $p(x_i\mid z_i)$ double-counts the unigram prior
of the token $x_i$, since both conditional distributions already contain the
intrinsic frequency bias of $x_i$. Without correction, the product would
over-emphasize high-frequency tokens. Therefore, the unigram prior $p(x_i)$
should be divided out.

Second, the residual term
\begin{equation}
    R(x_i;y,z_i)
    =
    \frac{
        p(y,z_i \mid x_i)
    }{
        p(y \mid x_i)\,p(z_i \mid x_i)
    }
\end{equation}
captures the interaction between the long context $y$ and the local prefix
$z_i$ that cannot be modeled by two independent conditional experts. Fully
modeling this residual may improve the approximation quality, but doing so
requires joint access to the target model, the parallel draft backbone, and the
causal correction module. This corresponds to the design philosophy of methods
such as Domino and DSpark, which attempt to learn such residual interactions
through end-to-end training.

In contrast, DeLS-Spec deliberately ignores this residual term in exchange for
modularity and training efficiency. We approximate
\begin{equation}
    \log p(x_i\mid y,z_i)
    \approx
    \log p_L(x_i\mid y)
    +
    \log p_S(x_i\mid z_i)
    -
    \log p_P(x_i),
\end{equation}
where $p_L$ is the long-context distribution provided by the trained
DFlash model, $p_S$ is an independently trained short-context model, and $p_P$
is the unigram prior estimated from the training corpus.

\subsection{Independent Short-Context Training}
\label{sec:short_training}

A key advantage of DeLS-Spec is that the short-context model can be trained
completely independently. It does not require hidden states from the target
model, nor does it require access to the DFlash draft model. The only
requirement is that it shares the same tokenizer as the target model. In our
implementation, $p_S(x_i\mid z_i)$ is parameterized by a lightweight local RNN
head and trained on plain text corpora with the standard next-token prediction
objective:
\begin{equation}
    \mathcal{L}_S
    =
    -\sum_{i=k+1}^{k+s}
    \log p_S(x_i \mid z_i),
    \qquad
    z_i=x_{k:i-1}.
\end{equation}
This training objective encourages the short-context model to capture local
causal dependencies inside a draft block, independent of the long-context
drafting model.

\subsection{Logit Fusion at Inference}
\label{sec:logit_fusion}

During inference, DeLS-Spec is directly attached to an already trained DFlash
draft model. Let $\ell_L(x_i\mid y)$ denote the long logits produced by DFlash,
$\ell_S(x_i\mid z_i)$ denote the short logits produced by the local RNN head,
and $\ell_P(x_i)$ denote the log unigram prior estimated from the training
corpus. We fuse these logits as
\begin{equation}
    \ell(x_i)
    =
    \ell_L(x_i\mid y)
    +
    \alpha\,\ell_S(x_i\mid z_i)
    -
    \beta\,\ell_P(x_i),
\end{equation}
where $\alpha$ and $\beta$ are calibration coefficients. The theoretical
decomposition corresponds to $\alpha=\beta=1$. In practice, however, the long
logits, short logits, and unigram prior are obtained from different sources and
may have different scales. Therefore, $\alpha$ and $\beta$ are introduced to
calibrate their relative contributions. The unigram prior term $\ell_P(x_i)$ is computed once from corpus statistics
and introduces no additional inference-time model computation, its estimation
details are provided in Appendix~\ref{app:implementation_details}. The
complete inference procedure is summarized in
Figure~\ref{fig:main-overview} and Algorithm~\ref{alg:dels_inference}.

\begin{algorithm}[t]
    \caption{DeLS-Spec Inference}
    \label{alg:dels_inference}
    \begin{algorithmic}[1]
        \Require Target model $M_T$, DFlash draft model $M_L$, local head $M_S$, unigram-prior logits $\ell_P$, fusion weights $\alpha,\beta$, draft span $s$, initial prefix $y$
        \Ensure Generated continuation appended to $y$
        \While{not finished}
            \State Get DFlash logits: $\{\ell_L^{(j)}\}_{j=0}^{s} \gets M_L(y)$.
            \State Sample the first draft token: $x_0 \sim \operatorname{Decode}(\operatorname{softmax}(\ell_L^{(0)}))$.
            \For{$j=1$ to $s$}
                \State Get short-context logits: $\ell_S^{(j)} \gets M_S(x_{0:j-1})$.
                \State Fuse logits: $\ell^{(j)} = \ell_L^{(j)} + \alpha\,\ell_S^{(j)} - \beta\,\ell_P$.
                \State Sample draft token: $x_j \sim \operatorname{Decode}(\operatorname{softmax}(\ell^{(j)}))$.
            \EndFor
            \State Verify $(x_0,\ldots,x_s)$ with one target-model forward pass.
            \State Append accepted tokens and the bonus token, if any, to $y$.
        \EndWhile
        \State \Return $y$
    \end{algorithmic}
\end{algorithm}

\section{Experiments}

\subsection{Experimental Setup}
\label{sec:experimental_setup}

\paragraph{Models and Benchmarks.}
We evaluate DeLS-Spec on Qwen3-4B and Qwen3-8B \citep{yang2025qwen3}. For the DFlash backbone \citep{chen2026dflash}, we use
the officially released checkpoints \texttt{z-lab/Qwen3-4B-DFlash-b16} and
\texttt{z-lab/Qwen3-8B-DFlash-b16}, both with a draft block size of 16. We
evaluate on math reasoning, code generation, and open-ended dialogue
benchmarks. The math benchmarks include GSM8K \citep{cobbe2021training}, MATH-500 \citep{hendrycks2021measuring}, and AIME25 \citep{aime25}; the code
benchmarks include HumanEval \citep{chen2021evaluating}, MBPP \citep{austin2021program}, and LiveCodeBench \citep{jain2025livecodebench}; and the dialogue
benchmarks include MT-Bench \citep{zheng2023judging} and Alpaca \citep{taori2023alpaca}. We report end-to-end decoding speedup
over autoregressive decoding and the average acceptance length $\tau$.

\paragraph{Training Data.}
We use the DFlash authors' \texttt{Qwen3-4B-Instruct-100K} data \citep{chen2026dflash} for the 4B
setting and the Domino authors' \texttt{Qwen3-8B-ShareGPT} data \citep{huang2026domino} for the 8B
setting. These corpora are used both to train the local head and to estimate the
unigram prior. Unlike methods that jointly fine-tune the draft pipeline,
DeLS-Spec only trains the local head with the standard next-token prediction
objective on the training corpus. The DFlash backbone used during evaluation is
kept fixed and directly loaded from the released \texttt{z-lab} checkpoints.

\paragraph{Baselines.}
We compare DeLS-Spec with autoregressive decoding and representative
speculative decoding baselines, including EAGLE-3, DART, and DFlash \citep{li2026eagle,liu2026dart,chen2026dflash}.
DFlash is the block-parallel backbone on which DeLS-Spec is
built. We also include additional comparisons with Domino-style fine-tuning and
other DFlash settings where applicable.

\paragraph{Implementation.}
Unless otherwise specified, all results are evaluated with the Hugging Face
Transformers backend \citep{wolf2019huggingface}. All training and evaluation runs are conducted on a
single NVIDIA L20 GPU. The local head is implemented as a GRU-based RNN by
default. For DeLS-Spec inference, we set
$\alpha=\beta=0.3$ unless otherwise stated. Following Domino, we optimize the
inference implementation with CUDA Graphs and fused Triton kernels \citep{tillet2019triton} to reduce
the overhead of the local-head decoding loop. Detailed local-head training
hyperparameters are provided in Table~\ref{tab:local-head-hparams}.

\subsection{Main Results}
\label{sec:main_results}

Table~\ref{tab:main-results} presents the main results on Qwen3-4B and Qwen3-8B 
across math, code, and chat benchmarks. DFlash already outperforms tree-based 
baselines such as EAGLE-3 and DART by a large margin. Built upon this strong 
baseline, DeLS-Spec further improves both decoding speedup and average 
acceptance length $\tau$ in almost all settings.

The gains are particularly clear on math and code tasks, where local token 
dependencies are stronger. For Qwen3-4B at temperature $0$, DeLS-Spec improves 
the speedup of DFlash from $4.74\times$ to $5.02\times$ on MBPP, 
from $6.09\times$ to $6.35\times$ on MATH-500, and from $5.69\times$ 
to $5.95\times$ on AIME25. Meanwhile, the acceptance length increases by 
up to $0.44$ on AIME25 and $0.40$ on MATH-500. Under stochastic decoding, 
DeLS-Spec also brings notable improvements, e.g., on Qwen3-4B at 
temperature $1$, HumanEval speedup increases from $4.61\times$ 
to $4.85\times$, with $\tau$ improved from $5.84$ to $6.21$. 
Similar trends hold for Qwen3-8B, where DeLS-Spec improves $\tau$ 
by $0.33$ on HumanEval and MBPP at temperature $0$, and by $0.29$ 
on MT-Bench at temperature $1$.

These results show that DeLS-Spec complements DFlash by 
improving the local consistency of draft tokens. The local head enables 
more drafted tokens to be accepted during verification, especially on 
math and code benchmarks, while preserving the original DFlash parallel 
drafting process.

\begin{table}[t]
    \caption{Decoding speedup over vanilla autoregressive decoding and average acceptance length ($\tau$) on Qwen3 models with a maximum of 2048 generated tokens. Parenthesized values indicate the draft tree size for EAGLE-3 and DART, and the draft block size for DFlash and DeLS-Spec. The average is computed over all listed benchmarks.}
    \label{tab:main-results}
    \resizebox{\linewidth}{!}{
        \scriptsize
        \centering
        \setlength{\tabcolsep}{1.2pt}
        \begin{tabular}{c l @{\hspace{1.0em}} cc cc cc @{\hspace{1.0em}} cc cc cc @{\hspace{1.0em}} cc cc @{\hspace{1.0em}} cc}
            \toprule
            \multirow{2}[2]{*}{Model} & \multirow{2}[2]{*}{\quad Method}
            & \multicolumn{6}{c@{\hspace{1.0em}}}{\sc{Math}}
            & \multicolumn{6}{c@{\hspace{1.0em}}}{\sc{Code}}
            & \multicolumn{4}{c@{\hspace{1.0em}}}{\sc{Chat}}
            & \multicolumn{2}{c}{\sc{Overall}} \\
            \cmidrule(lr){3-8}
            \cmidrule(lr){9-14}
            \cmidrule(lr){15-18}
            \cmidrule(lr){19-20}
            & & \multicolumn{2}{c}{GSM8K}
            & \multicolumn{2}{c}{MATH-500}
            & \multicolumn{2}{c@{\hspace{1.0em}}}{AIME25}
            & \multicolumn{2}{c}{HumanEval}
            & \multicolumn{2}{c}{MBPP}
            & \multicolumn{2}{c@{\hspace{1.0em}}}{LCB}
            & \multicolumn{2}{c}{MT-Bench}
            & \multicolumn{2}{c@{\hspace{1.0em}}}{Alpaca}
            & \multicolumn{2}{c}{\textit{Avg.}} \\
            \midrule

            \multicolumn{2}{c}{Temperature = 0}
            & \tabsmall Speedup & \tabsmall $\tau$
            & \tabsmall Speedup & \tabsmall $\tau$
            & \tabsmall Speedup & \tabsmall $\tau$
            & \tabsmall Speedup & \tabsmall $\tau$
            & \tabsmall Speedup & \tabsmall $\tau$
            & \tabsmall Speedup & \tabsmall $\tau$
            & \tabsmall Speedup & \tabsmall $\tau$
            & \tabsmall Speedup & \tabsmall $\tau$
            & \tabsmall Speedup & \tabsmall $\tau$ \\

            \midrule
            \multirow{5}{*}{4B}
            & \mbox{EAGLE-3~\tabsmall\secondarycolor{(16)}}
            & 2.24$\times$ & 3.32
            & 2.10$\times$ & 3.11
            & 2.08$\times$ & 3.10
            & 2.09$\times$ & 3.09
            & 2.02$\times$ & 2.99
            & 1.95$\times$ & 2.93
            & 1.95$\times$ & 2.94
            & 1.87$\times$ & 2.83
            & 2.04$\times$ & 3.04 \\

            & \mbox{EAGLE-3~\tabsmall\secondarycolor{(60)}}
            & 2.57$\times$ & 3.83
            & 2.40$\times$ & 3.59
            & 2.36$\times$ & 3.53
            & 2.36$\times$ & 3.53
            & 2.30$\times$ & 3.44
            & 2.19$\times$ & 3.31
            & 2.25$\times$ & 3.41
            & 2.14$\times$ & 3.29
            & 2.32$\times$ & 3.49 \\

            & \mbox{DART~\tabsmall\secondarycolor{(60)}}
            & 2.16$\times$ & 2.65
            & 2.20$\times$ & 2.63
            & 2.13$\times$ & 2.59
            & 2.48$\times$ & 2.99
            & 2.47$\times$ & 3.01
            & 2.23$\times$ & 2.76
            & 2.19$\times$ & 2.69
            & 2.16$\times$ & 2.78
            & 2.25$\times$ & 2.76 \\

            & DFlash~\tabsmall\secondarycolor{(16)}
            & 5.15$\times$ & 6.51
            & 6.09$\times$ & 7.81
            & 5.69$\times$ & 7.32
            & 5.19$\times$ & 6.56
            & 4.74$\times$ & 6.14
            & 5.25$\times$ & 6.82
            & 2.72$\times$ & 4.13
            & 2.23$\times$ & 3.06
            & 4.63$\times$ & 6.04 \\

            & DeLS-Spec~\tabsmall\secondarycolor{(16)}
            & \textbf{5.29$\times$} & \textbf{6.80}
            & \textbf{6.35$\times$} & \textbf{8.21}
            & \textbf{5.95$\times$} & \textbf{7.76}
            & \textbf{5.42$\times$} & \textbf{6.92}
            & \textbf{5.02$\times$} & \textbf{6.47}
            & \textbf{5.50$\times$} & \textbf{7.20}
            & \textbf{2.77$\times$} & \textbf{4.30}
            & \textbf{2.25$\times$} & \textbf{3.15}
            & \textbf{4.82$\times$} & \textbf{6.35} \\

            \midrule
            \multirow{5}{*}{8B}
            & \mbox{EAGLE-3~\tabsmall\secondarycolor{(16)}}
            & 2.21$\times$ & 3.27
            & 2.09$\times$ & 3.10
            & 2.07$\times$ & 3.07
            & 2.17$\times$ & 3.21
            & 1.93$\times$ & 2.86
            & 1.80$\times$ & 2.95
            & 1.82$\times$ & 2.75
            & 1.67$\times$ & 2.53
            & 1.97$\times$ & 2.97 \\

            & \mbox{EAGLE-3~\tabsmall\secondarycolor{(60)}}
            & 2.56$\times$ & 3.80
            & 2.42$\times$ & 3.61
            & 2.41$\times$ & 3.59
            & 2.50$\times$ & 3.74
            & 2.22$\times$ & 3.31
            & 2.03$\times$ & 3.12
            & 2.07$\times$ & 3.17
            & 1.88$\times$ & 2.90
            & 2.26$\times$ & 3.41 \\

            & \mbox{DART~\tabsmall\secondarycolor{(60)}}
            & 2.28$\times$ & 2.71
            & 2.29$\times$ & 2.70
            & 2.11$\times$ & 2.59
            & 2.52$\times$ & 2.95
            & 2.39$\times$ & 2.98
            & 2.24$\times$ & 2.78
            & 2.27$\times$ & 3.03
            & 2.21$\times$ & 2.84
            & 2.29$\times$ & 2.82 \\

            & DFlash~\tabsmall\secondarycolor{(16)}
            & 4.96$\times$ & 6.59
            & 5.83$\times$ & 7.85
            & 5.20$\times$ & 7.05
            & 4.98$\times$ & 6.60
            & 4.56$\times$ & 6.08
            & 5.15$\times$ & 7.17
            & 2.59$\times$ & 4.16
            & 2.16$\times$ & 3.11
            & 4.43$\times$ & 6.08 \\

            & DeLS-Spec~\tabsmall\secondarycolor{(16)}
            & \textbf{5.08$\times$} & \textbf{6.84}
            & \textbf{5.98$\times$} & \textbf{8.15}
            & \textbf{5.40$\times$} & \textbf{7.35}
            & \textbf{5.16$\times$} & \textbf{6.93}
            & \textbf{4.72$\times$} & \textbf{6.41}
            & \textbf{5.16$\times$} & \textbf{7.27}
            & \textbf{2.66$\times$} & \textbf{4.35}
            & \textbf{2.18$\times$} & \textbf{3.20}
            & \textbf{4.54$\times$} & \textbf{6.31} \\

            \midrule
            \multicolumn{2}{c}{Temperature = 1}
            & \tabsmall Speedup & \tabsmall $\tau$
            & \tabsmall Speedup & \tabsmall $\tau$
            & \tabsmall Speedup & \tabsmall $\tau$
            & \tabsmall Speedup & \tabsmall $\tau$
            & \tabsmall Speedup & \tabsmall $\tau$
            & \tabsmall Speedup & \tabsmall $\tau$
            & \tabsmall Speedup & \tabsmall $\tau$
            & \tabsmall Speedup & \tabsmall $\tau$
            & \tabsmall Speedup & \tabsmall $\tau$ \\

            \midrule
            \multirow{5}{*}{4B}
            & \mbox{EAGLE-3~\tabsmall\secondarycolor{(16)}}
            & 2.18$\times$ & 3.26
            & 2.00$\times$ & 3.00
            & 1.86$\times$ & 2.80
            & 2.02$\times$ & 3.04
            & 1.97$\times$ & 2.95
            & 1.87$\times$ & 2.83
            & 1.90$\times$ & 2.91
            & 1.78$\times$ & 2.72
            & 1.95$\times$ & 2.94 \\

            & \mbox{EAGLE-3~\tabsmall\secondarycolor{(60)}}
            & 2.44$\times$ & 3.77
            & 2.28$\times$ & 3.51
            & 2.08$\times$ & 3.10
            & 2.09$\times$ & 3.09
            & 2.19$\times$ & 3.38
            & 2.05$\times$ & 3.19
            & 2.11$\times$ & 3.33
            & 2.02$\times$ & 3.17
            & 2.16$\times$ & 3.32 \\

            & \mbox{DART~\tabsmall\secondarycolor{(60)}}
            & 2.19$\times$ & 2.65
            & 2.20$\times$ & 2.63
            & 2.13$\times$ & 2.59
            & 2.48$\times$ & 2.99
            & 2.46$\times$ & 3.01
            & 2.23$\times$ & 2.76
            & 2.19$\times$ & 2.69
            & 2.16$\times$ & 2.78
            & 2.26$\times$ & 2.76 \\

            & DFlash~\tabsmall\secondarycolor{(16)}
            & 4.69$\times$ & 5.95
            & 5.02$\times$ & 6.59
            & 3.69$\times$ & 4.90
            & 4.61$\times$ & 5.84
            & 4.43$\times$ & 5.68
            & 5.06$\times$ & 6.66
            & \textbf{2.61$\times$} & 3.89
            & 2.12$\times$ & 2.90
            & 4.03$\times$ & 5.30 \\

            & DeLS-Spec~\tabsmall\secondarycolor{(16)}
            & \textbf{4.82$\times$} & \textbf{6.22}
            & \textbf{5.14$\times$} & \textbf{6.99}
            & \textbf{3.81$\times$} & \textbf{5.04}
            & \textbf{4.85$\times$} & \textbf{6.21}
            & \textbf{4.58$\times$} & \textbf{5.89}
            & \textbf{5.15$\times$} & \textbf{6.79}
            & 2.59$\times$ & \textbf{3.94}
            & \textbf{2.16$\times$} & \textbf{3.00}
            & \textbf{4.14$\times$} & \textbf{5.51} \\

            \midrule
            \multirow{5}{*}{8B}
            & \mbox{EAGLE-3~\tabsmall\secondarycolor{(16)}}
            & 2.18$\times$ & 3.26
            & 1.96$\times$ & 3.00
            & 1.86$\times$ & 2.80
            & 2.05$\times$ & 3.09
            & 1.97$\times$ & 2.95
            & 1.90$\times$ & 2.93
            & 1.91$\times$ & 2.91
            & 1.78$\times$ & 2.83
            & 1.95$\times$ & 2.97 \\

            & \mbox{EAGLE-3~\tabsmall\secondarycolor{(60)}}
            & 2.40$\times$ & 3.70
            & 2.23$\times$ & 3.44
            & 2.05$\times$ & 3.17
            & 2.32$\times$ & 3.58
            & 2.11$\times$ & 3.24
            & 1.90$\times$ & 2.95
            & 1.91$\times$ & 3.02
            & 1.78$\times$ & 2.81
            & 2.09$\times$ & 3.24 \\

            & \mbox{DART~\tabsmall\secondarycolor{(60)}}
            & 2.25$\times$ & 2.71
            & 2.25$\times$ & 2.70
            & 2.13$\times$ & 2.59
            & 2.45$\times$ & 2.95
            & 2.45$\times$ & 2.98
            & 2.25$\times$ & 2.78
            & 2.24$\times$ & 2.73
            & 2.18$\times$ & 2.84
            & 2.28$\times$ & 2.79 \\

            & DFlash~\tabsmall\secondarycolor{(16)}
            & 4.40$\times$ & 5.88
            & 4.69$\times$ & 6.48
            & 3.27$\times$ & 4.57
            & 4.13$\times$ & 5.48
            & 4.01$\times$ & 5.36
            & 4.84$\times$ & 6.79
            & 2.38$\times$ & 3.73
            & 2.03$\times$ & 2.93
            & 3.72$\times$ & 5.15 \\

            & DeLS-Spec~\tabsmall\secondarycolor{(16)}
            & \textbf{4.51$\times$} & \textbf{6.09}
            & \textbf{4.78$\times$} & \textbf{6.73}
            & \textbf{3.36$\times$} & \textbf{4.82}
            & \textbf{4.26$\times$} & \textbf{5.75}
            & \textbf{4.12$\times$} & \textbf{5.64}
            & \textbf{4.96$\times$} & \textbf{6.96}
            & \textbf{2.44$\times$} & \textbf{4.02}
            & \textbf{2.04$\times$} & \textbf{2.96}
            & \textbf{3.81$\times$} & \textbf{5.37} \\

            \bottomrule
        \end{tabular}
    }
\end{table}

\subsection{General Adaptability of the DeLS-Spec Local Head}
\label{sec:local_head_adaptability}

We further test the general adaptability of the local head on
two DFlash block-7 checkpoints from DSpark
\citep{dspark}: \texttt{deepseek-ai/dflash\_qwen3\_4b\_block7} and
\texttt{deepseek-ai/dflash\_qwen3\_8b\_block7}.
This setting examines whether the local head can transfer beyond our default
DFlash configuration. Since the local head is trained with block size $16$,
it can be directly used with DFlash models whose block size is no larger than
$16$.

As shown in Table~\ref{tab:dspark-dflash-block7-results}, DeLS-Spec 
consistently improves these DSpark checkpoints. On the 4B checkpoint, 
DeLS-Spec brings speedup gains on MATH-500, AIME25, 
and HumanEval, improving DFlash by $0.26\times$ on each benchmark. 
It also increases the acceptance length by $0.33$ on HumanEval and $0.31$ 
on MATH-500. On the 8B checkpoint, DeLS-Spec improves $\tau$ by $0.32$ on 
GSM8K and MATH-500, and by $0.31$ on LCB, while also increasing speedup 
across all benchmarks.

These consistent improvements on independently released block-7 checkpoints 
demonstrate that the DeLS-Spec local head is not tied to a specific DFlash 
implementation. Instead, it can serve as a plug-in module that improves 
local draft-token modeling for different DFlash-style draft models.

\begin{table}[t]
    \caption{Results of directly applying DeLS-Spec to DSpark's DFlash block-7 checkpoints. DeLS-Spec achieves consistent performance improvements across all benchmarks.}
    \label{tab:dspark-dflash-block7-results}
    \resizebox{\linewidth}{!}{
        \scriptsize
        \centering
        \setlength{\tabcolsep}{2.0pt}
        \begin{tabular}{c l @{\hspace{1.0em}} cc cc cc @{\hspace{1.0em}} cc cc cc @{\hspace{1.0em}} cc cc @{\hspace{1.0em}} cc}
            \toprule
            \multirow{2}[2]{*}{Model} & \multirow{2}[2]{*}{\quad Method}
            & \multicolumn{2}{c}{GSM8K}
            & \multicolumn{2}{c}{MATH-500}
            & \multicolumn{2}{c@{\hspace{1.0em}}}{AIME25}
            & \multicolumn{2}{c}{HumanEval}
            & \multicolumn{2}{c}{MBPP}
            & \multicolumn{2}{c@{\hspace{1.0em}}}{LCB}
            & \multicolumn{2}{c}{MT-Bench}
            & \multicolumn{2}{c@{\hspace{1.0em}}}{Alpaca}
            & \multicolumn{2}{c}{\textit{Avg.}} \\
            \cmidrule(lr){3-4}
            \cmidrule(lr){5-6}
            \cmidrule(lr){7-8}
            \cmidrule(lr){9-10}
            \cmidrule(lr){11-12}
            \cmidrule(lr){13-14}
            \cmidrule(lr){15-16}
            \cmidrule(lr){17-18}
            \cmidrule(lr){19-20}
            \multicolumn{2}{c}{}
            & \tabsmall Speedup & \tabsmall $\tau$
            & \tabsmall Speedup & \tabsmall $\tau$
            & \tabsmall Speedup & \tabsmall $\tau$
            & \tabsmall Speedup & \tabsmall $\tau$
            & \tabsmall Speedup & \tabsmall $\tau$
            & \tabsmall Speedup & \tabsmall $\tau$
            & \tabsmall Speedup & \tabsmall $\tau$
            & \tabsmall Speedup & \tabsmall $\tau$
            & \tabsmall Speedup & \tabsmall $\tau$ \\
            \midrule
            \multirow{2}{*}{4B}
            & DFlash
            & 4.03$\times$ & 4.93
            & 3.79$\times$ & 4.65
            & 3.21$\times$ & 3.93
            & 3.10$\times$ & 3.75
            & 3.38$\times$ & 4.13
            & 3.25$\times$ & 3.96
            & 2.43$\times$ & 3.17
            & 2.26$\times$ & 2.87
            & 3.18$\times$ & 3.92 \\

            & DeLS-Spec
            & \textbf{4.24$\times$} & \textbf{5.20}
            & \textbf{4.05$\times$} & \textbf{4.96}
            & \textbf{3.47$\times$} & \textbf{4.21}
            & \textbf{3.36$\times$} & \textbf{4.08}
            & \textbf{3.59$\times$} & \textbf{4.42}
            & \textbf{3.44$\times$} & \textbf{4.23}
            & \textbf{2.53$\times$} & \textbf{3.37}
            & \textbf{2.34$\times$} & \textbf{3.00}
            & \textbf{3.38$\times$} & \textbf{4.18} \\

            \midrule
            \multirow{2}{*}{8B}
            & DFlash
            & 4.05$\times$ & 4.84
            & 3.73$\times$ & 4.49
            & 3.29$\times$ & 3.87
            & 3.10$\times$ & 3.68
            & 3.44$\times$ & 4.13
            & 3.38$\times$ & 4.04
            & 2.49$\times$ & 3.23
            & 2.34$\times$ & 2.91
            & 3.23$\times$ & 3.90 \\

            & DeLS-Spec
            & \textbf{4.21$\times$} & \textbf{5.16}
            & \textbf{3.91$\times$} & \textbf{4.81}
            & \textbf{3.36$\times$} & \textbf{4.08}
            & \textbf{3.26$\times$} & \textbf{3.97}
            & \textbf{3.55$\times$} & \textbf{4.34}
            & \textbf{3.53$\times$} & \textbf{4.35}
            & \textbf{2.56$\times$} & \textbf{3.39}
            & \textbf{2.40$\times$} & \textbf{3.05}
            & \textbf{3.35$\times$} & \textbf{4.14} \\
            \bottomrule
        \end{tabular}
    }
\end{table}

\subsection{Effect of \texorpdfstring{$\alpha$ and $\beta$}{alpha and beta}}
\label{sec:alpha-beta-scan}

\begin{figure}[t]
    \centering
    \begin{minipage}{0.32\linewidth}
        \centering
        \includegraphics[width=\linewidth]{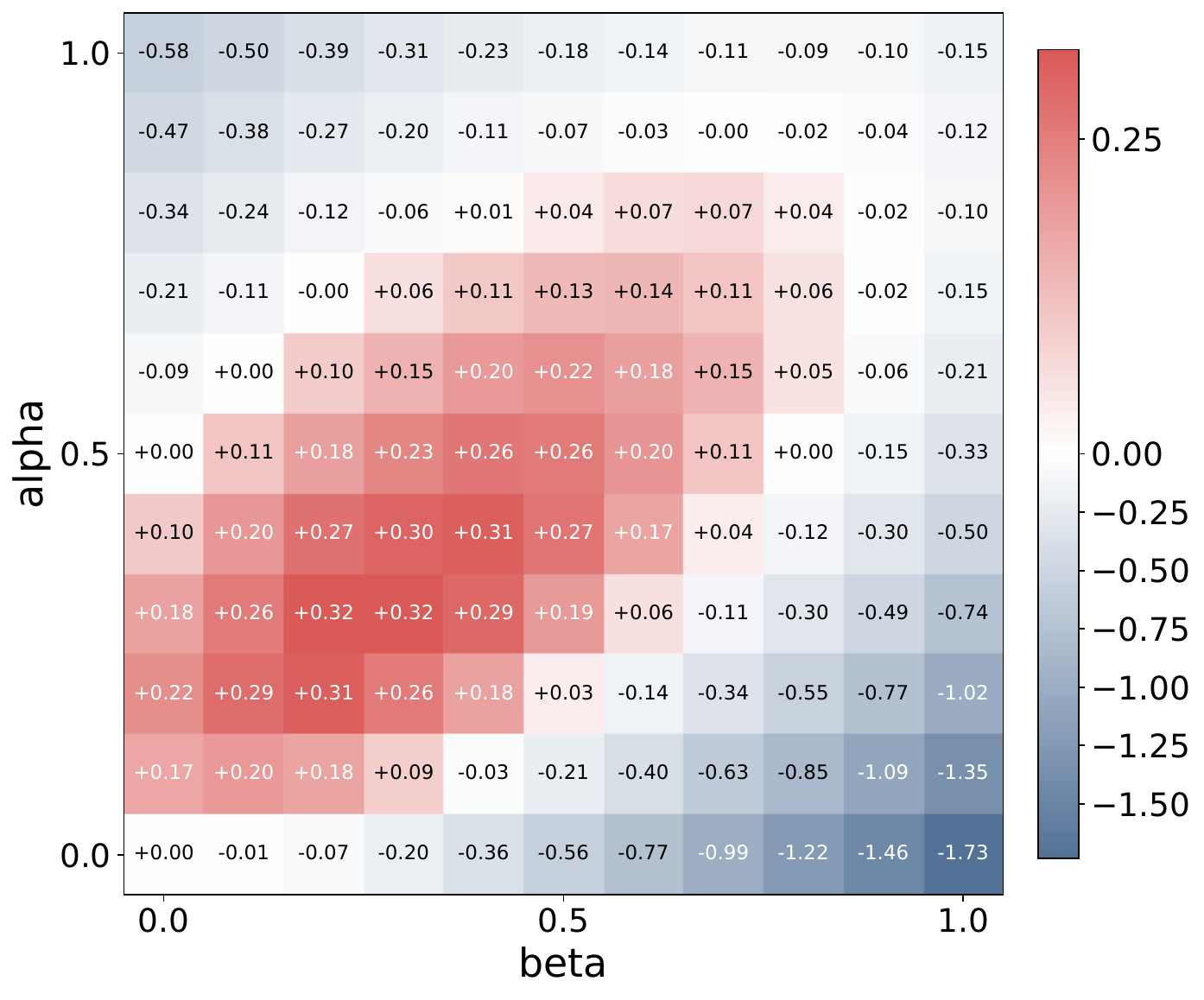}
    \end{minipage}\hfill
    \begin{minipage}{0.32\linewidth}
        \centering
        \includegraphics[width=\linewidth]{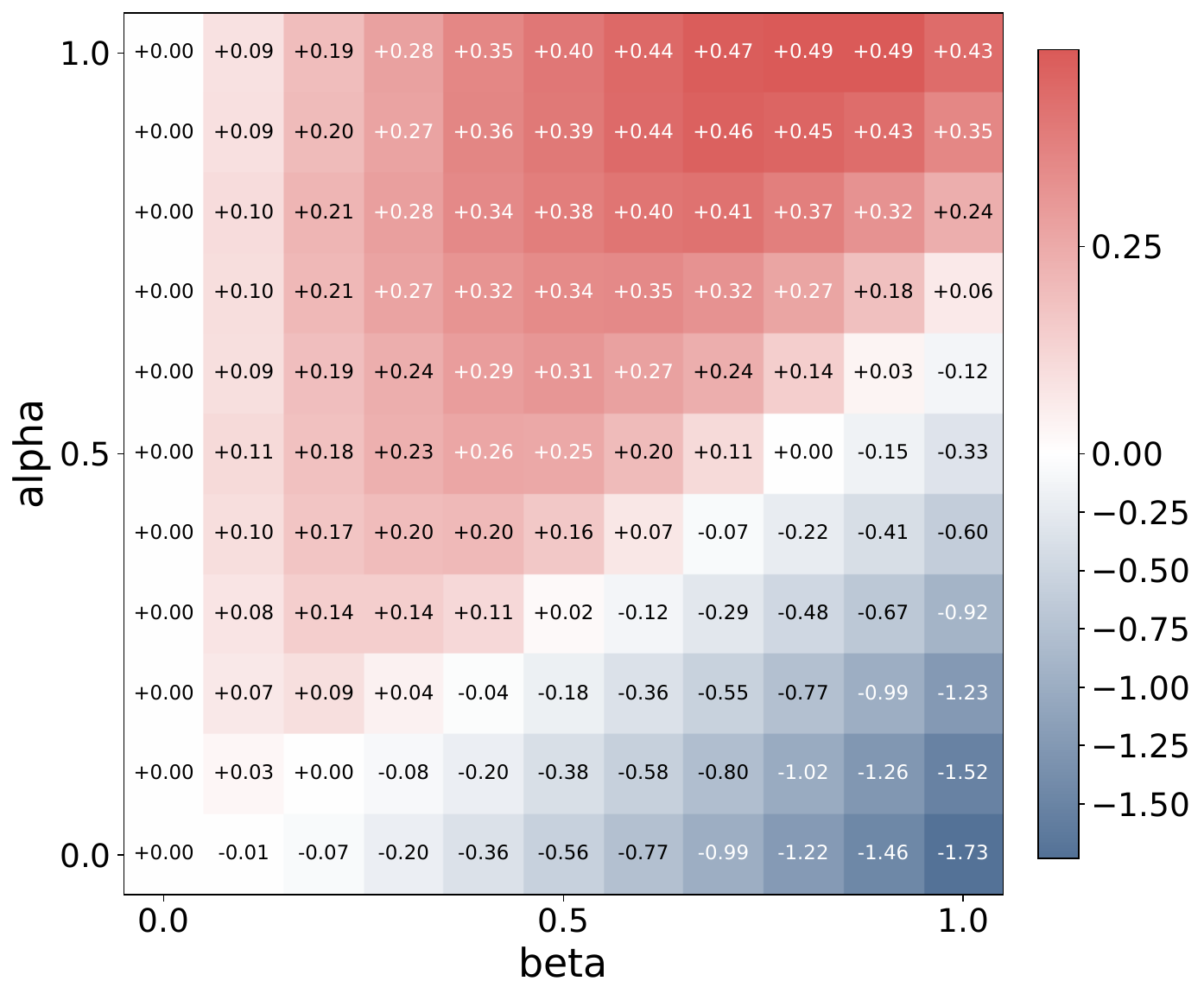}
    \end{minipage}\hfill
    \begin{minipage}{0.32\linewidth}
        \centering
        \includegraphics[width=\linewidth]{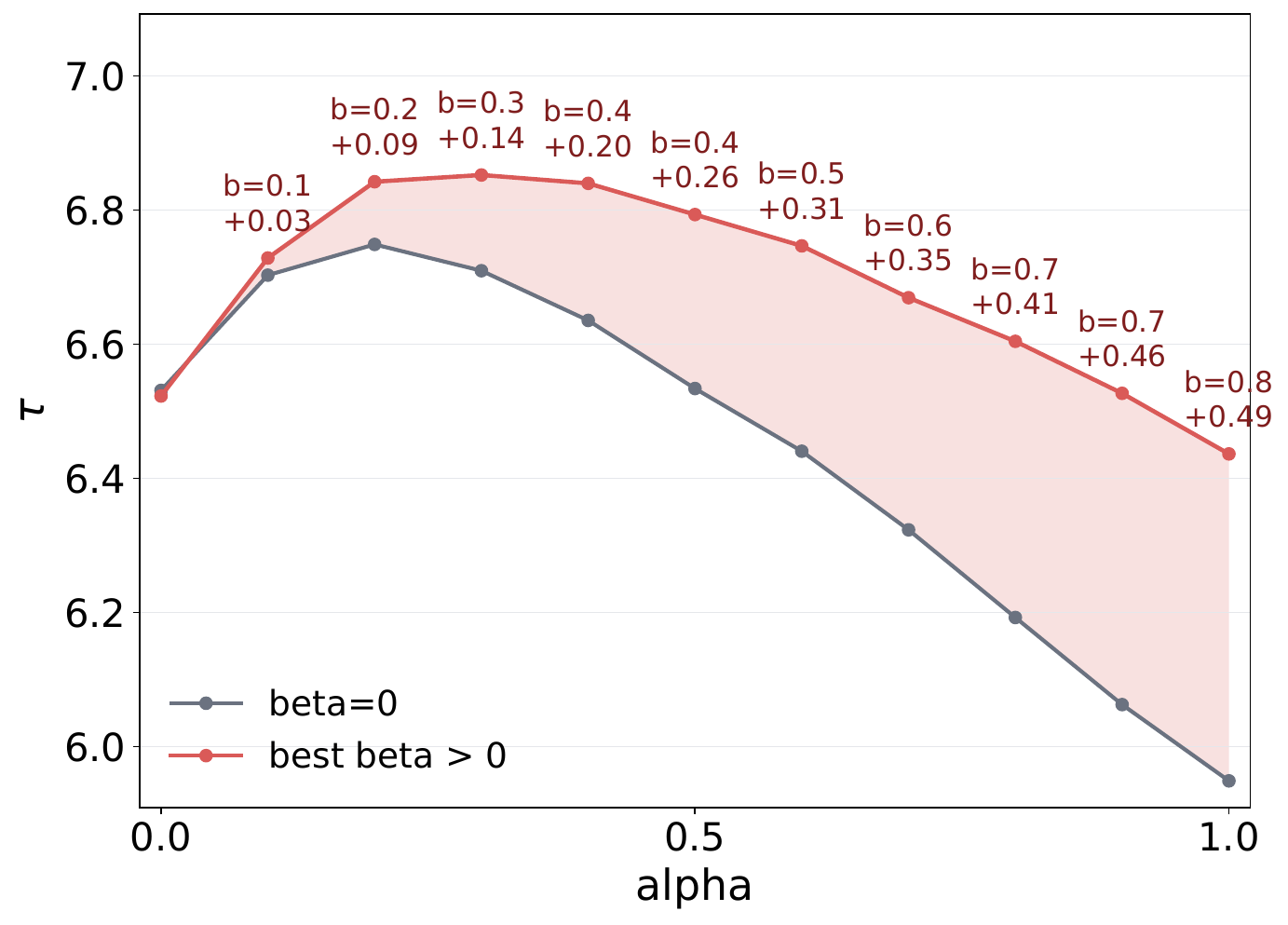}
    \end{minipage}
    \caption{$\alpha , \; \beta$ scan on Qwen3-4B. Left: average acceptance length under different $\alpha$ and $\beta$ values. Middle: row-wise improvement after subtracting the corresponding $\beta=0$ value. Right: improvement of the best $\beta$ for each $\alpha$ over $\beta=0$.}
    \label{fig:scan-results}
\end{figure}

We further study the effect of the two combination hyperparameters, $\alpha$ and $\beta$, by performing a grid search on Qwen3-4B. Figure~\ref{fig:scan-results} summarizes the results. The left panel shows that $\alpha$ and $\beta$ should be increased jointly to achieve the best performance, rather than tuning either one alone. The heatmap also exhibits a clear diagonal trend, suggesting that in practice $\alpha$ and $\beta$ can simply be set to the same value, with values below $0.5$ yielding the best performance. In particular, the best region lies around $\alpha \approx 0.3$ and $\beta \approx 0.3$, indicating that a balanced contribution from the local head and the unigram-prior correction is important.

To better highlight the importance of subtracting the unigram prior, the middle panel normalizes each row by subtracting the corresponding value at $\beta=0$. This row-wise comparison makes the effect of $\beta$ more explicit: once $\alpha$ is properly set, using a positive $\beta$ consistently improves the acceptance length over the $\beta=0$ case. The right panel further summarizes this trend by plotting, for each $\alpha$, the improvement of the best $\beta$ over $\beta=0$. We observe consistent gains across almost all $\alpha$ values, which confirms that unigram-prior subtraction is a key component of the proposed formulation.

\subsection{Impact of the Omitted Residual Term}
\label{sec:residual-term-impact}

We next quantify how much performance is lost by omitting the residual term $R(x_i; y, z_i)$ in DeLS-Spec. Table~\ref{tab:compare-results} compares several variants in terms of average acceptance length $\tau$. \textit{DeLS-Spec (Markov)} follows the Markov head used in DSpark \citep{dspark}, where the logits of each draft token are corrected only based on the immediately preceding token. In our framework, this corresponds to generating the short-context logits using only the previous token. \textit{Domino-FT} freezes the DFlash backbone and trains only the Domino head \citep{huang2026domino}, using the same training data and number of epochs as our DeLS-Spec local head.

Although Domino-FT achieves the highest average $\tau$, the gap between DeLS-Spec and Domino-FT is small. On average, DFlash obtains $\tau=6.04$, DeLS-Spec (RNN) improves it to $6.35$, while Domino-FT reaches $6.45$. Thus, omitting the residual term only leads to a $0.10$ drop in average acceptance length compared with Domino-FT, while DeLS-Spec still captures $75.69\%$ of Domino-FT's improvement over DFlash. The $\tau$ gain ratio is computed as
\(
\frac{\tau_{\text{DeLS-Spec}}-\tau_{\text{DFlash}}}
{\tau_{\text{Domino-FT}}-\tau_{\text{DFlash}}}.
\)

The gains are also consistent across benchmarks. DeLS-Spec (RNN) recovers more than $80\%$ of Domino-FT's improvement on AIME25 and HumanEval, and even surpasses Domino-FT on LCB with a $\tau$ gain ratio of $105.56\%$. In addition, DeLS-Spec (RNN) consistently outperforms DeLS-Spec (Markov), improving the average $\tau$ from $6.28$ to $6.35$. This shows that modeling richer short-range dependencies with an RNN local head is more effective than a one-step Markov correction, while still avoiding the full residual modeling cost of Domino-FT.

\begin{table}[t]
    \caption{
Average acceptance length comparison for estimating the impact of the omitted
residual term. The gain ratio reports how much of the Domino-FT improvement over
DFlash is recovered by DeLS-Spec without explicitly learning the residual term.
}
    \label{tab:compare-results}
    \resizebox{\linewidth}{!}{
        \scriptsize
        \centering
        \setlength{\tabcolsep}{3.0pt}
        \begin{tabular}{l c c c c c c c c c}
            \toprule
            Method & GSM8K & MATH-500 & AIME25 & HumanEval & MBPP & LCB & MT-Bench & Alpaca & Avg. \\
            \midrule
            DFlash & 6.51 & 7.81 & 7.32 & 6.56 & 6.14 & 6.82 & 4.13 & 3.06 & 6.04 \\
            DeLS-Spec (Markov) & 6.69 & 8.17 & 7.69 & 6.82 & 6.38 & 7.09 & 4.29 & 3.12 & 6.28 \\
            DeLS-Spec (RNN) & 6.80 & 8.21 & 7.76 & 6.92 & 6.47 & 7.20 & 4.30 & 3.15 & 6.35 \\
            Domino-FT & 7.03 & 8.37 & 7.85 & 7.01 & 6.60 & 7.18 & 4.37 & 3.19 & 6.45 \\
            $\tau$ gain ratio & 55.77\% & 71.43\% & 83.02\% & 80.00\% & 71.74\% & 105.56\% & 70.83\% & 69.23\% & 75.69\% \\
            \bottomrule
        \end{tabular}
    }
\end{table}

\subsection{Do Learnable \texorpdfstring{$\alpha$ and $\beta$}{alpha and beta} Help?}
\label{sec:learnable-alpha-beta}

We further investigate whether learning the fusion weights $\alpha$ and $\beta$ can bring additional gains over fixed values. As shown in Table~\ref{tab:learnable-results}, $\alpha=\beta=0$ corresponds to the original DFlash baseline, while $\alpha=\beta=0.3$ is our default setting. For the learnable variant, we initialize $\alpha$ and $\beta$ to $0.3$ and optimize them for 10K steps on the training data. During this process, the target model, DFlash, and the local head are all loaded but frozen, and only $\alpha$ and $\beta$ are updated with the cross-entropy loss.

The learnable variant consistently improves over the DFlash baseline, increasing the average acceptance length from $6.33$ to $6.55$. This confirms that assigning positive weights to the local correction and unigram-prior subtraction is beneficial. However, it is still worse than our simple fixed setting $\alpha=\beta=0.3$, which achieves the best average $\tau$ of $6.65$ and outperforms the learnable variant on every benchmark. For example, the fixed setting improves $\tau$ over the learnable variant by $0.20$ on MATH-500, $0.13$ on GSM8K and MBPP, and $0.06$ on LCB.

Appendix~\ref{app:learnable-alpha-beta-analysis} reports the learned values of
$\alpha$ and $\beta$ in Table~\ref{tab:learnable-alpha-beta-value}. They are
consistently positive and increase with draft-token position, which supports the
usefulness of both terms. Nevertheless, directly optimizing these weights with
cross-entropy does not translate to better acceptance length on downstream
benchmarks. This suggests that the fixed $\alpha=\beta=0.3$ setting provides a
better trade-off between effectiveness, robustness, and simplicity, avoiding the
need to load all components and perform extra tuning.

\begin{table}[t]
\caption{
Comparison between fixed and learnable fusion weights on Qwen3-4B.
The fixed setting $\alpha=\beta=0.3$ provides the best average acceptance
length, while learnable weights offer limited additional benefit and can
slightly underperform the tuned fixed fusion.
}
    \label{tab:learnable-results}
    \resizebox{\linewidth}{!}{
        \scriptsize
        \centering
        \setlength{\tabcolsep}{10.0pt}
        \begin{tabular}{l c c c c c c c}
            \toprule
            $\alpha \; \& \; \beta$ & GSM8K & MATH-500 & HumanEval & MBPP & LCB & MT-Bench & Avg. \\
            \midrule
            $\alpha=\beta=0$ & 6.51 & 7.81 & 6.56 & 6.14 & 6.82 & 4.13 & 6.33 \\
            $\alpha=\beta=0.3$ & 6.80 & 8.21 & 6.92 & 6.47 & 7.20 & 4.30 & 6.65 \\
            Learnable & 6.67 & 8.01 & 6.89 & 6.34 & 7.15 & 4.24 & 6.55 \\
            \bottomrule
        \end{tabular}
    }
\end{table}

\subsection{Training Cost Comparison}
\label{sec:training-cost-comparison}

We compare the training cost of Domino-FT and DeLS-Spec on a single NVIDIA L20 GPU with 48GB memory. As shown in Table~\ref{tab:training-cost-results}, all methods are trained with batch size $1$, accumulation steps $4$, and one epoch. For Domino-FT and DeLS-Spec (RNN), we sample 256 anchors from each example to construct draft blocks. In contrast, DeLS-Spec (Markov) does not rely on anchors or block construction, and is trained with dense token-level supervision directly on the corpus.

\begin{table}[t]
\caption{
Training cost comparison on a single NVIDIA L20 GPU (48G).
DeLS-Spec local heads require much less training time and memory than
Domino-FT, with the Markov variant being the cheapest due to dense supervision
and its minimal parameterization.
}
    \label{tab:training-cost-results}
    \centering
    \resizebox{\linewidth}{!}{
        \scriptsize
        \setlength{\tabcolsep}{10.0pt}
        \begin{tabular}{c l l c c}
            \toprule
            Model & Method & Supervision & Training time (h) & VRAM (GB) \\
            \midrule
            \multirow{3}{*}{Qwen3-4B}
            & Domino-FT & anchor / block & 13.4 & 42.6 \\
            & DeLS-Spec (RNN) & anchor / block & 1.1 & 9.0 \\
            & DeLS-Spec (Markov) & dense supervision & 0.4 & 6.5 \\
            \midrule
            \multirow{3}{*}{Qwen3-8B}
            & Domino-FT & anchor / block & N/A & OOM \\
            & DeLS-Spec (RNN) & anchor / block & 1.1 & 10.1 \\
            & DeLS-Spec (Markov) & dense supervision & 0.5 & 5.9 \\
            \bottomrule
        \end{tabular}
    }
\end{table}

DeLS-Spec is substantially more efficient than Domino-FT. On Qwen3-4B, Domino-FT requires 13.4 hours and 42.6GB VRAM, while DeLS-Spec (RNN) only takes 1.1 hours and 9.0GB VRAM, reducing training time by over $12\times$ and memory usage by about $4.7\times$. The Markov variant is even cheaper, requiring only 0.4 hours and 6.5GB VRAM. The gap becomes more pronounced on Qwen3-8B: Domino-FT runs out of memory on a 48GB L20 GPU, whereas DeLS-Spec (RNN) still fits comfortably with 10.1GB VRAM and finishes training in 1.1 hours. DeLS-Spec (Markov) further reduces the cost to 0.5 hours and 5.9GB VRAM.

These results highlight the practical advantage of DeLS-Spec. The key reason is that the DeLS-Spec local head can be trained independently of both the target model and DFlash. In contrast, Domino-FT requires the target model to produce hidden states and DFlash to provide base logits, while the Domino head only learns the residual correction logits. As a result, Domino-FT must keep multiple large components involved during training, leading to much higher memory usage and training time. DeLS-Spec avoids this dependency by training a lightweight local head separately, making it significantly easier to apply under limited hardware resources.

\section{Conclusion}

We presented \textbf{DeLS-Spec}, a decoupled long-short context method for block-parallel speculative decoding. DeLS-Spec keeps DFlash fixed as a long-context expert and adds a lightweight local head as a short-context expert. By combining their logits at inference time, it introduces intra-block causal information without retraining the draft model from scratch.
The local head can be independently trained with a standard next-token 
prediction objective, making training cheap and modular. 
Since it is not one-to-one tied to a specific DFlash checkpoint, 
it can also be flexibly attached to compatible DFlash-style drafters. 
Experiments on Qwen3 models show that DeLS-Spec consistently improves speedup and average acceptance length over DFlash, demonstrating an efficient way to enhance existing block-parallel speculative decoding systems.

\section{Limitations and Future Work}

While DeLS-Spec demonstrates significant efficiency and performance gains, our approach has several limitations that present opportunities for future research:

\textbf{Omission of Residual Interactions.} To achieve extreme training efficiency and modularity, DeLS-Spec deliberately omits the residual interaction term between the long context and the local short context. As a result, while it recovers the majority of performance improvements, its average acceptance length is slightly bounded and falls marginally short of end-to-end jointly trained methods like Domino, which explicitly learn this full residual interaction.

\textbf{Generalization to Other Parallel Drafters.} Our current empirical evaluation is primarily built upon the DFlash architecture. However, the theoretical framework of DeLS-Spec is fundamentally agnostic to the underlying draft model. Since the local head operates independently, future work will explore applying and validating DeLS-Spec on a broader range of non-autoregressive and parallel draft models beyond DFlash.

\textbf{Adaptive Logit Fusion Strategies.} Currently, DeLS-Spec employs a straightforward logit fusion mechanism with fixed hyperparameters ($\alpha$ and $\beta$). We believe the fusion process can be further optimized. Future research could explore dynamic fusion strategies, such as adaptively adjusting the local head and unigram-prior weights based on the token-level entropy or confidence scores of the long-context logits, allowing for a more context-aware balance between global semantics and local causality.

\textbf{Mitigating Exposure Mismatch in Local Heads.} As observed in our analysis, directly learning the fusion weights ($\alpha$ and $\beta$) via cross-entropy yields suboptimal inference performance compared to using fixed values. We attribute this to an exposure mismatch (or teacher-forcing bias) during training, where the local head is conditioned on ground-truth prefixes rather than its own drafted tokens. Future work could investigate techniques such as scheduled sampling or alignment-based tuning to bridge this train-test gap, potentially unlocking the full capability of learnable fusion parameters.

\newpage
\bibliography{iclr2026_conference}
\bibliographystyle{iclr2026_conference}

\newpage
\appendix

\section{Implementation Details}
\label{app:implementation_details}

\paragraph{Local Head Implementation.}
We instantiate $p_S(x_i\mid z_i)$ with a lightweight local head over the local
draft prefix. Since the local head is evaluated autoregressively during
inference, we use a low-rank bottleneck to reduce the sequential computation
cost, following the design choice in Domino and DSpark \citep{huang2026domino,dspark}. For the
RNN variant, token ids are mapped by the frozen target embedding table and
passed through a one-layer bias-free GRU; the final hidden state is projected to
a low-rank state with SiLU activation and then mapped to vocabulary logits:
\[
    r_i=\mathrm{SiLU}\!\left(W_{\mathrm{rank}}
    \mathrm{GRU}(E_{\mathrm{tgt}}(z_i))\right),
    \qquad
    \ell_S^{\mathrm{RNN}}(x_i\mid z_i)=W_{\mathrm{vocab}}r_i .
\]
For the Markov variant, we directly look up a low-rank state from the most
recent token and multiply it with a low-rank vocabulary head:
\[
    r_i=E_M(x_{i-1}),
    \qquad
    \ell_S^{\mathrm{Markov}}(x_i\mid z_i)=r_i W_M^\top ,
\]
where $E_M,W_M\in\mathbb{R}^{|\mathcal{V}|\times d_r}$. Both variants output
full-vocabulary logits and are trained with the same next-token prediction
objective as the short-context model. The hyperparameters for the local head 
are summarized in Table~\ref{tab:local-head-hparams}.

\paragraph{Unigram Prior Estimation.}
We construct a fixed unigram log-prior over target tokens using the same
tokenizer and chat-template preprocessing as training. For each training
sequence, we count only tokens whose loss mask equals one, i.e.,
assistant-response tokens that contribute to the supervised objective. Let
$c(v)$ denote the resulting count for vocabulary item $v$. We form a smoothed
prior
\[
    \tilde p_0(v)
    =
    \frac{c(v)+\lambda}{\sum_{v'}(c(v')+\lambda)},
\]
where we use additive smoothing with $\lambda=1.0$ by default. We then use
$b_0(v)=\log \tilde p_0(v)$ as a fixed vocabulary-level bias.

\begin{table}[H]
    \caption{Training hyperparameters for the local head.}
    \label{tab:local-head-hparams}
    \centering
    \setlength{\tabcolsep}{18.0pt}
    \begin{tabular}{l c}
        \toprule
        Hyperparameter & Value \\
        \midrule
        block-size & 16 \\
        num-anchors & 512 \\
        local-gru-hidden-dim & 1024 \\
        local-rank & 256 \\
        local-rank-activation & SiLU \\
        loss-decay-gamma & 7.0 \\
        max-length & 3072 \\
        batch-size & 4 \\
        accumulation-steps & 1 \\
        learning-rate & 6e-4 \\
        num-epochs & 1 \\
        \bottomrule
    \end{tabular}
\end{table}


\newpage

\section{Proof of the Product-of-Experts Factorization}
\label{app:proof_factorization}

In Section~\ref{sec:dels}, we introduce the exact factorization of the conditional probability $p(x_i \mid y, z_i)$. Here, we provide the step-by-step derivation for this decomposition.

Starting from Bayes' theorem, the conditional probability of the candidate token $x_i$ given the long context $y$ and the short context $z_i$ can be written as:
\begin{equation}
    p(x_i \mid y, z_i) = \frac{p(y, z_i \mid x_i)\,p(x_i)}{p(y, z_i)}.
\end{equation}

We can multiply both the numerator and the denominator by the product of the marginals $p(y)\,p(z_i)$ and the conditional probabilities $p(y \mid x_i)\,p(z_i \mid x_i)$:
\begin{equation}
    p(x_i \mid y, z_i) = \frac{p(y, z_i \mid x_i)\,p(x_i)}{p(y, z_i)} \cdot \frac{p(y \mid x_i)\,p(z_i \mid x_i)}{p(y \mid x_i)\,p(z_i \mid x_i)} \cdot \frac{p(y)\,p(z_i)}{p(y)\,p(z_i)}.
\end{equation}

By rearranging the terms, we group them into three distinct factors:
\begin{equation}
    p(x_i \mid y, z_i) = \left[ \frac{p(y \mid x_i)\,p(z_i \mid x_i)\,p(x_i)}{p(y)\,p(z_i)} \right] \cdot \frac{p(y)\,p(z_i)}{p(y, z_i)} \cdot \frac{p(y, z_i \mid x_i)}{p(y \mid x_i)\,p(z_i \mid x_i)}.
    \label{eq:proof_rearranged}
\end{equation}

Recall that by applying Bayes' theorem to the individual contexts, we can express the conditional probabilities of $x_i$ as:
\begin{equation}
    p(x_i \mid y) = \frac{p(y \mid x_i)\,p(x_i)}{p(y)} \quad \text{and} \quad p(x_i \mid z_i) = \frac{p(z_i \mid x_i)\,p(x_i)}{p(z_i)}.
\end{equation}

Multiplying these two equations together and dividing by the unigram prior $p(x_i)$ yields the exact expression inside the square brackets of Equation~\ref{eq:proof_rearranged}:
\begin{equation}
    \frac{p(x_i \mid y)\,p(x_i \mid z_i)}{p(x_i)} = \frac{p(y \mid x_i)\,p(z_i \mid x_i)\,p(x_i)}{p(y)\,p(z_i)}.
\end{equation}

Substituting this equivalence back into Equation~\ref{eq:proof_rearranged} gives the final exact equality:
\begin{equation}
\begin{aligned}
    p(x_i \mid y, z_i)
    &=
    \frac{
        p(x_i \mid y)\,p(x_i \mid z_i)
    }{
        p(x_i)
    }
    \cdot
    \frac{
        p(y)\,p(z_i)
    }{
        p(y, z_i)
    }
    \cdot
    \frac{
        p(y, z_i \mid x_i)
    }{
        p(y \mid x_i)\,p(z_i \mid x_i)
    } .
\end{aligned}
\end{equation}
This completes the proof.

\newpage
\section{Analysis of Learnable \texorpdfstring{$\alpha$ and $\beta$}{alpha and beta}}
\label{app:learnable-alpha-beta-analysis}

Table~\ref{tab:learnable-alpha-beta-value} reports the learned values of $\alpha$ and $\beta$ at different draft-token positions. We observe that both weights become relatively large at later positions. For example, $\alpha$ increases from $0.547$ at position $2$ to around $0.762$ after position $12$, while $\beta$ also increases from $0.512$ to around $0.613$. This trend suggests that the learned fusion weights assign increasingly higher importance to the local head and the unigram-prior correction for later draft tokens.

A possible explanation is the mismatch introduced by teacher forcing during training \citep{bengio2015scheduled}. 
When optimizing $\alpha$ and $\beta$ on the training data, later positions are conditioned on longer ground-truth prefixes. Under this setting, the local head can exploit more accurate short-context information and may appear more reliable than DFlash for predicting later draft tokens. As a result, the optimization tends to assign larger $\alpha$ and $\beta$ values to later positions.

However, this advantage may not fully transfer to inference. During speculative decoding, later draft tokens are conditioned on previously drafted tokens rather than ground-truth tokens, which introduces exposure bias. Therefore, overly large learned weights at later positions may overestimate the reliability of the local correction and lead to suboptimal acceptance length. This explains why the learnable variant in Table~\ref{tab:learnable-results} improves over the DFlash baseline but still underperforms the fixed setting $\alpha=\beta=0.3$. The fixed setting is less sensitive to the teacher-forcing bias and provides a more robust choice for inference.

\begin{table}[H]
    \caption{Learnable $\alpha$ and $\beta$ values by position.}
    \label{tab:learnable-alpha-beta-value}
    \centering
    \setlength{\tabcolsep}{10.0pt}
    \begin{tabular}{c c c}
        \toprule
        Position & \(\alpha\) & \(\beta\) \\
        \midrule
        2 & 0.547 & 0.512 \\
        3 & 0.594 & 0.523 \\
        4 & 0.633 & 0.539 \\
        5 & 0.672 & 0.559 \\
        6 & 0.699 & 0.578 \\
        7 & 0.715 & 0.586 \\
        8 & 0.730 & 0.598 \\
        9 & 0.742 & 0.605 \\
        10 & 0.750 & 0.613 \\
        11 & 0.754 & 0.613 \\
        12 & 0.762 & 0.613 \\
        13 & 0.762 & 0.617 \\
        14 & 0.762 & 0.613 \\
        15 & 0.762 & 0.613 \\
        \bottomrule
    \end{tabular}
\end{table}

\newpage
\section{Comparison of Acceptance Rate Results}
\label{app:acceptance-rate-results}

Table~\ref{tab:acceptance-rate-results} provides a position-wise view of the
acceptance behavior on Qwen3-4B at temperature $0$. Unlike the average
acceptance length $\tau$ reported in Table~\ref{tab:main-results}, the
quantity $rate_i=P(\mathrm{acceptance\ length}\ge i)$ measures the probability
that verification accepts at least the first $i$ draft tokens. It therefore
describes the tail of the accepted draft length distribution: improvements at
larger positions indicate that the method more often preserves longer
consecutive draft spans, which is the direct source of higher average
acceptance length and decoding speedup.

The results show that DeLS-Spec improves the acceptance curve over DFlash
across almost all nontrivial positions. The first position is accepted by both
methods by definition, and the second position is nearly unchanged
($81.45\%$ versus $81.63\%$). From position $3$ onward, however, DeLS-Spec
consistently achieves higher acceptance rates. The gains are largest in the
middle of the draft block, reaching about $2.6$ percentage points around
positions $5$--$6$, and remain positive even at the end of the block, with a
$1.40$ percentage-point improvement at position $16$. This pattern supports the
main hypothesis of DeLS-Spec: the local head and unigram-prior correction do
not merely improve isolated early predictions, but make the draft sequence
locally more coherent so that verification can accept longer prefixes.

\begin{table}[H]
    \caption{Acceptance-rate comparison by position on Qwen3-4B at temperature 0, where \(rate_i = P(\mathrm{acceptance\ length} \ge i)\).}
    \label{tab:acceptance-rate-results}
    \centering
    \setlength{\tabcolsep}{8.0pt}
    \begin{tabular}{c c c c}
        \toprule
        Position & DFlash & DeLS-Spec & DeLS-Spec - DFlash \\
        \midrule
        1 & 100.00\% & 100.00\% & +0.00\% \\
        2 & 81.63\% & 81.45\% & -0.18\% \\
        3 & 63.98\% & 65.10\% & +1.13\% \\
        4 & 50.65\% & 52.69\% & +2.03\% \\
        5 & 41.26\% & 43.80\% & +2.54\% \\
        6 & 34.38\% & 36.99\% & +2.61\% \\
        7 & 29.26\% & 31.63\% & +2.37\% \\
        8 & 25.15\% & 27.49\% & +2.34\% \\
        9 & 22.02\% & 24.11\% & +2.10\% \\
        10 & 19.45\% & 21.24\% & +1.79\% \\
        11 & 17.14\% & 18.85\% & +1.71\% \\
        12 & 15.26\% & 16.83\% & +1.57\% \\
        13 & 13.44\% & 14.95\% & +1.51\% \\
        14 & 11.96\% & 13.42\% & +1.46\% \\
        15 & 10.51\% & 11.96\% & +1.45\% \\
        16 & 9.22\% & 10.62\% & +1.40\% \\
        \bottomrule
    \end{tabular}
\end{table}

\end{document}